\documentclass[10pt,twocolumn,letterpaper]{article}

\usepackage[pagenumbers]{iccv} 

\definecolor{iccvblue}{rgb}{0.21,0.49,0.74}
\usepackage[pagebackref,breaklinks,colorlinks,allcolors=iccvblue]{hyperref}

\usepackage[accsupp]{axessibility}
\usepackage{algorithm}
\usepackage{algorithmicx}
\usepackage{algpseudocode}
\usepackage{multirow}

\usepackage{xcolor}
\usepackage[theorems,skins]{tcolorbox}
\usepackage{empheq}
\tcbset{red eqbox/.style={enhanced,top=-0.1ex, bottom=-0.1ex, 
left=-0.2ex,right=-0.2ex,
overlay={\fill[red!5] (frame.south west) to[bend left] 
 (frame.north west) --  (frame.north east) to[bend left]
  (frame.south east) -- cycle;},
boxrule=0pt},
red1 eqbox/.style={enhanced,top=-0.1ex, bottom=-0.1ex, 
left=-0.2ex,right=-0.2ex,
overlay={\fill[red!10] (frame.south west) to[bend left] 
 (frame.north west) --  (frame.north east) to[bend left]
  (frame.south east) -- cycle;},
boxrule=0pt},
red2 eqbox/.style={enhanced,top=-0.1ex, bottom=-0.1ex, 
left=-0.2ex,right=-0.2ex,
overlay={\fill[red!15] (frame.south west) to[bend left] 
 (frame.north west) --  (frame.north east) to[bend left]
  (frame.south east) -- cycle;},
boxrule=0pt},
red3 eqbox/.style={enhanced,top=-0.1ex, bottom=-0.1ex, 
left=-0.2ex,right=-0.2ex,
overlay={\fill[red!20] (frame.south west) to[bend left] 
 (frame.north west) --  (frame.north east) to[bend left]
  (frame.south east) -- cycle;},
boxrule=0pt},
red4 eqbox/.style={enhanced,top=-0.1ex, bottom=-0.1ex, 
left=-0.2ex,right=-0.2ex,
overlay={\fill[red!25] (frame.south west) to[bend left] 
 (frame.north west) --  (frame.north east) to[bend left]
  (frame.south east) -- cycle;},
boxrule=0pt},
red5 eqbox/.style={enhanced,top=-0.1ex, bottom=-0.1ex, 
left=-0.2ex,right=-0.2ex,
overlay={\fill[red!30] (frame.south west) to[bend left] 
 (frame.north west) --  (frame.north east) to[bend left]
  (frame.south east) -- cycle;},
boxrule=0pt},
blue eqbox/.style={enhanced,top=-0.1ex, bottom=-0.1ex, 
left=-0.2ex,right=-0.2ex,
overlay={\fill[blue!10] (frame.south west) to[bend left] 
 (frame.north west) --  (frame.north east) to[bend left]
  (frame.south east) -- cycle;},
boxrule=0pt},
blue1 eqbox/.style={enhanced,top=-0.1ex, bottom=-0.1ex, 
left=-0.2ex,right=-0.2ex,
overlay={\fill[blue!5] (frame.south west) to[bend left] 
 (frame.north west) --  (frame.north east) to[bend left]
  (frame.south east) -- cycle;},
boxrule=0pt},
blue2 eqbox/.style={enhanced,top=-0.1ex, bottom=-0.1ex, 
left=-0.2ex,right=-0.2ex,
overlay={\fill[blue!10] (frame.south west) to[bend left] 
 (frame.north west) --  (frame.north east) to[bend left]
  (frame.south east) -- cycle;},
boxrule=0pt},
blue3 eqbox/.style={enhanced,top=-0.1ex, bottom=-0.1ex, 
left=-0.2ex,right=-0.2ex,
overlay={\fill[blue!15] (frame.south west) to[bend left] 
 (frame.north west) --  (frame.north east) to[bend left]
  (frame.south east) -- cycle;},
boxrule=0pt},
blue4 eqbox/.style={enhanced,top=-0.1ex, bottom=-0.1ex, 
left=-0.2ex,right=-0.2ex,
overlay={\fill[blue!20] (frame.south west) to[bend left] 
 (frame.north west) --  (frame.north east) to[bend left]
  (frame.south east) -- cycle;},
boxrule=0pt},
blue5 eqbox/.style={enhanced,top=-0.1ex, bottom=-0.1ex, 
left=-0.2ex,right=-0.2ex,
overlay={\fill[blue!25] (frame.south west) to[bend left] 
 (frame.north west) --  (frame.north east) to[bend left]
  (frame.south east) -- cycle;},
boxrule=0pt},
blue6 eqbox/.style={enhanced,top=-0.1ex, bottom=-0.1ex, 
left=-0.2ex,right=-0.2ex,
overlay={\fill[blue!30] (frame.south west) to[bend left] 
 (frame.north west) --  (frame.north east) to[bend left]
  (frame.south east) -- cycle;},
boxrule=0pt},
highlight math style=red eqbox,
}
\definecolor{BlueViolet}{rgb}{0.54, 0.17, 0.89}
\definecolor{RedViolet}{rgb}{0.78, 0.08, 0.52}

\NewDocumentCommand{\customsubsubsection}{m}{\noindent\textbf{#1. }}

\def\paperID{13083} 
\def\confName{ICCV}
\def\confYear{2025}

\title{Draw Your Mind: Personalized Generation via Condition-Level Modeling in Text-to-Image Diffusion Models\vspace{-0.8em}}

\author{Hyungjin Kim, Seokho Ahn, Young-Duk Seo$^*$\\
Department of Electrical and Computer Engineering, Inha University\\
{\tt\small flslzk@inha.edu, sokho0514@inha.edu, mysid88@inha.ac.kr}
\\{\small $^{*}$Corresponding author}
\\{\small \textcolor{blue}{https://github.com/Burf/DrUM}}
\vspace{-0.2em}
}

\begin{document}

\twocolumn[{%
\maketitle
\vspace{-2.9em}
\renewcommand\twocolumn[1][]{#1}%
\begin{center}
    \centering
    \captionsetup{type=figure}
    \includegraphics[width=1\textwidth]{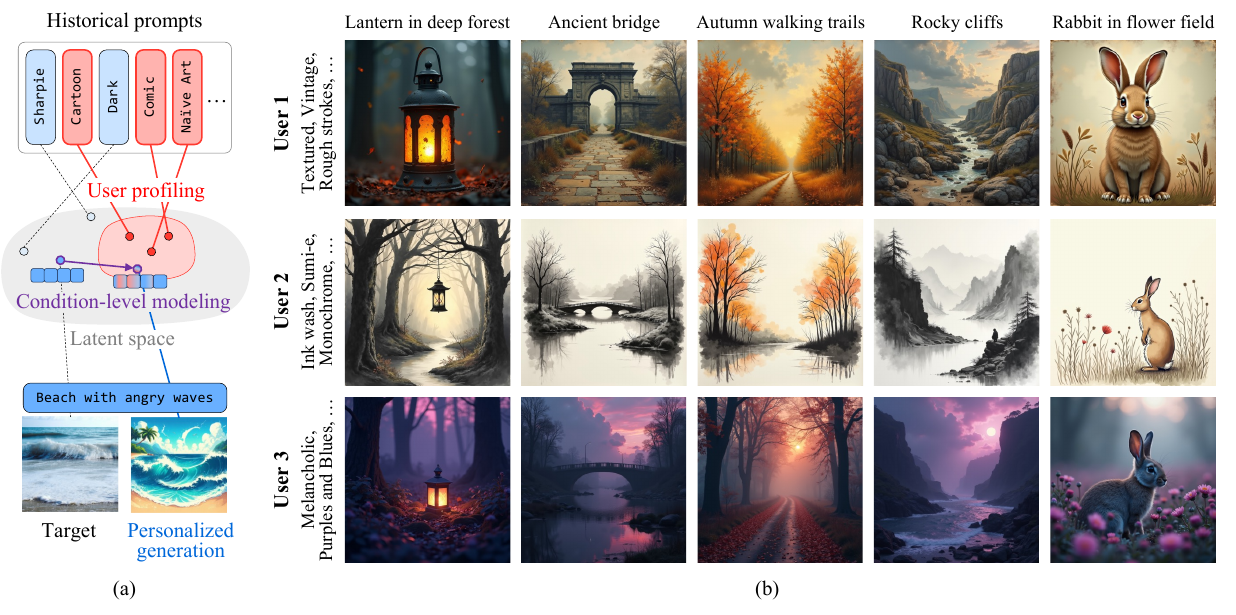}
    \vspace{-2em}
    \caption{DrUM personalizes synthesis results on popular foundation T2I models without fine-tuning by incorporating open-source text encoders. It integrates key preference information such as style, texture within the latent space, effectively satisfying individual demands. (a) Conceptual illustration of our methodology. (b) Personalized results using \textit{multiple prompts} on Stable Diffusion V3.}
    \label{fig_teaser}
\end{center}%
}]

\begin{abstract}
Personalized generation in T2I diffusion models aims to naturally incorporate individual user preferences into the generation process with minimal user intervention. However, existing studies primarily rely on prompt-level modeling with large-scale models, often leading to inaccurate personalization due to the limited input token capacity of T2I diffusion models. To address these limitations, we propose DrUM, a novel method that integrates user profiling with a transformer-based adapter to enable personalized generation through condition-level modeling in the latent space. DrUM demonstrates strong performance on large-scale datasets and seamlessly integrates with open-source text encoders, making it compatible with widely used foundation T2I models without requiring additional fine-tuning.
\end{abstract}    
\vspace{-3.6em}
\section{Introduction}
\label{sec:intro}
\vspace{0.7em}
Recently, various techniques in text-to-image (T2I) diffusion models, such as fine-tuning, additional conditioning, editing, and guidance, have been developed. These approaches enable users to exert more direct control over the generation process beyond prompt engineering, allowing for greater alignment with their intended outcomes \cite{huang2024diffusion}. However, they often require substantial computational resources and iterative refinements before achieving results that accurately reflect user preferences \cite{von2023fabric, salehi2025viper}.

Personalized generation presents a promising solution to these limitations by directly integrating individual preferences into the generation process \cite{von2023fabric, chen2024tailored, salehi2025viper, shen2024pmg, xu2025personalized}. In this context, many studies have employed large language models (LLMs) and multimodal models to generate personalized inputs for T2I models \cite{chen2024tailored, salehi2025viper, shen2024pmg, xu2025personalized}. However, these approaches primarily rely on prompt-level modeling, which is restricted by the limited input length and the constrained references imposed by the token capacity of text encoders. Furthermore, the dependence on large-scale models results in significant computational costs. To enable personalized generation with minimal user intervention while effectively addressing individual needs, the following challenges must be addressed: (i) accurately identifying individual preferences, (ii) overcoming the constraints of prompt-level modeling, and (iii) precisely incorporating user preferences into the generation process.

To address these challenges, we propose DrUM (\textbf{Dr}aw \textbf{You}r \textbf{M}ind), a novel approach to achieve all the core requirements for personalized generation. The DrUM first constructs a user profile by applying coreset sampling (\Cref{sec_coreset}) to extract key information from the user's history. It then employs a transformer-based adapter for condition-level modeling (\Cref{sec_peca}), enabling the integration of diverse user preferences into a personalized T2I model input within the latent space. Additionally, DrUM leverages a guidance mechanism (\Cref{sec_cond_guid}) within the adapter to precisely capture detailed user profiles while maintaining minimal structural complexity, offering flexible control over preference intensity and personalization degree. Notably, it seamlessly incorporates with open-source text encoders, such as OpenCLIP \cite{cherti2023reproducible} and Google T5 \cite{raffel2020exploring}, making it compatible with widely used foundation T2I models, including Stable Diffusion V1/V2/XL/V3 \cite{rombach2022high, podell2024sdxl, esser2024scaling} and FLUX, without requiring additional fine-tuning.

The main contributions of this study are as follows: 
\begin{itemize}
    \item We introduce DrUM, the first \textit{de facto} approach to personalized generation via condition-level modeling in T2I diffusion models.
    \item DrUM effectively integrates individual preferences into high-quality personalized synthesis results by considering both preference intensity and personalization degree, overcoming the limitations of prompt-level modeling. 
    \item DrUM achieves superior performance while preserving creativity and diversity across various foundation T2I models, all without requiring additional fine-tuning.
\end{itemize}

\section{Related works}
We review existing studies on T2I diffusion models and personalized generation.

\subsection{Text-to-image diffusion models}
T2I diffusion models primarily generate realistic images by using text prompts as input. GLIDE \cite{nichol2021glide} was the first study to produce high-dimensional pixel-level images from text by employing classifier-free guidance \cite{ho2022classifier}. DALL-E 2 \cite{ramesh2022hierarchical} adopted the OpenCLIP text encoder to generate images that accurately reflect user inputs. Imagen \cite{saharia2022photorealistic} utilized a hierarchical structure and the Google T5 encoder to create high-quality images that fully capture textual semantics. Stable Diffusion \cite{rombach2022high} introduced the latent diffusion model (LDM), a standard T2I architecture that leverages the OpenCLIP text encoder to perform diffusion in a lower-dimensional latent space. The most widely adopted T2I diffusion architectures are based on LDM with open-source text encoders to achieve precise generative capabilities. These models have significantly enhanced users' creative activities, and various methods have been explored to meet the increasing and diverse demands of users.

Fine-tuning methods like DreamBooth \cite{ruiz2023dreambooth} effectively reflect input images as a special token within personalized diffusion models. However, they still require substantial resources for additional training and rely on well-designed prompts to achieve satisfactory results. Additional condition methods like ControlNet \cite{zhang2023adding} allows users to directly control desired composition through additional inputs, but they struggle to fully capture semantic information such as style and texture. Editing methods like InstructPix2Pix \cite{brooks2023instructpix2pix} enable precise edits but often demand additional inputs or manual adjustments to produce the desired outcomes. Guidance methods like StyleDiffusion \cite{wang2023stylediffusion} generate results emphasizing specific attributes but require expert user intervention for fine adjustments. In general, when users aim for precise results with T2I diffusion models, a higher level of expertise and significant time investment are required. Moreover, the process may require additional input data or high-performance hardware \cite{huang2024diffusion}.

\subsection{Personalized generation}
\label{sec_rel_per}

Personalized generation aims to naturally incorporate individual user preferences into the generative process, providing highly satisfactory results with minimal user intervention. This approach has demonstrated its effectiveness across various domains, including fashion \cite{yu2019personalized,xu2024diffusion}, music \cite{dai2022personalised,plitsis2024investigating}, modeling \cite{chen2023train,li2024text}, and advertising \cite{vashishtha2024chaining,yang2024new}, and recent attempts have extended to diffusion models.

Fabric \cite{von2023fabric} utilizes attention-level guidance during the self-attention process of Stable Diffusion V1/V2 to reflect key image-level features, but it can only consider a single positive/negative image pair. Tailored Visions (TV) \cite{chen2024tailored} uses ChatGPT 3.5 to rewrite prompts by retrieving three historical prompts similar to the input using OpenCLIP ViT-L. However, considering more than three prompts degrades generation quality, and necessitates additional resources for API calls and retrieval. ViPer \cite{salehi2025viper} employs the multi-modal model IDEFICS2-8b \cite{laurenccon2024matters} to construct a single like/dislike prompt pair for virtual users' visual preferences and uses simple contrastive guidance to transform this prompt pair into personalized input on Stable Diffusion V1/XL. However, it relies only on a single pair within a limited input length and requires at least 8–20 images with detailed comments for user profiling. PMG \cite{shen2024pmg} employs LLaMA2-7B \cite{touvron2023llama} to reconstruct inputs using keywords extracted from users' preferred products, but it integrates only limited preference information and is restricted to specific domains such as movie posters and stickers. Additionally, it introduces soft preference embedding to complement keyword modeling, but this focuses on considering the next preferred product rather than reflecting the user's overall preferences. Pigeon \cite{xu2025personalized} fine-tunes the multi-modal generative model LaVIT \cite{jin2024unified} for personalized generation using both historical prompts and images. However, it relies on LaVIT's built-in LDM and remains limited to specific domains such as movie posters and stickers. Additionally, it requires reference images for personalized generation, but maintaining large amounts of these images incurs significant costs, making scalability a challenge.
\section{Preliminaries}
This section describes preliminaries on T2I conditioning and personalized generation in T2I diffusion models.

\subsection{Text encoder-based T2I conditioning}
T2I conditioning is the process that text encoders transform prompts into conditions for T2I generation \cite{huang2024diffusion}. Recent trends have shifted the focus from smaller encoders like OpenCLIP ViT-L to more powerful architectures such as OpenCLIP ViT-bigG and Google T5. These larger-scale encoders with optimized combinations have been shown to enhance the quality of T2I conditioning, leading to more semantically accurate outputs \cite{podell2024sdxl, esser2024scaling}.

Notably, the T2I condition $y$ directly influences the variation in generated results, as modeled in the reverse diffusion process with the following formula:
\begin{equation}
    p_\theta(x_{t-1}|x_t, y) = \mathcal{N}\Bigl(x_{t-1}; \mu_\theta(x_t, t, y), \Sigma_\theta(x_t, t, y)\Bigr)
\end{equation}
Here, the conditional probability $p_\theta(x_{t-1} | x_t, y)$ models the transition from the current state $x_t$ to the previous state $x_{t-1}$ under the T2I condition $y$, with mean $\mu_\theta$ and covariance $\Sigma_\theta$ parameterizing the Gaussian distribution to guide denoising for semantically faithful image generation.

Despite the effectiveness of this approach, the unique combinations of various text encoders in each T2I model make it challenging to establish a unified method for consistently improving all models.

\subsection{Personalized generation in T2I diffusion models}
\label{sec_pre_per}

As introduced in the previous section, personalized generation identifies user preferences as a profile and integrates this profile into the generation process to provide personalized content. However, existing studies on three key steps of personalized generation have several limitations:
\begin{itemize}
    \item \textbf{User profiling}: Identifying individual preferences is a crucial step in personalized generation \cite{salehi2025viper}, but it only leverages a limited number of historical entries \cite{chen2024tailored, shen2024pmg, xu2025personalized} or relies on detailed user feedback \cite{von2023fabric, salehi2025viper}.
    \item \textbf{Personalized T2I input}: Existing methods primarily model T2I inputs at the prompt-level, which limits expressiveness due to restricted input token length \cite{chen2024tailored, salehi2025viper, shen2024pmg}. Additionally, incorporating more preference information often degrades performance \cite{chen2024tailored, shen2024pmg}.
    \item \textbf{Integrating user preference}: Precise personalization requires fine-grained user profile integration during the generation process. However, most studies rely only on large-scale models, even for detailed information (\ie, preference intensity and personalization degree) \cite{chen2024tailored, salehi2025viper, shen2024pmg, xu2025personalized}.
\end{itemize}
In summary, achieving personalized generation that accurately reflects individual preferences with minimal user intervention is crucial. This requires a design that ensures both creativity and diversity for a wide range of users while addressing these limitations.
\section{Methodology}
This section introduces DrUM, a method designed to sample key user preferences and generate precise synthesis results accordingly.

\begin{figure*}[t]
\centering
\includegraphics[width=1\textwidth]{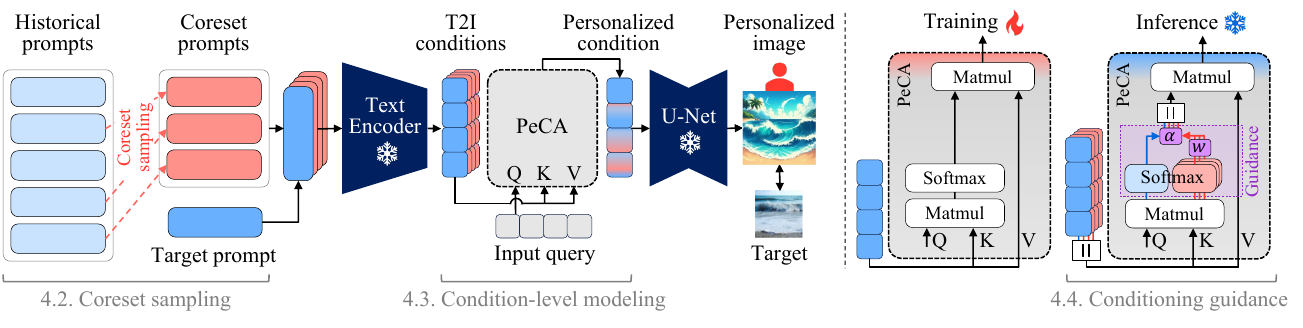}
\vspace{-1.7em}
\caption{Overview of methodology. \(||\) denotes concatenation.}
\label{fig_methodology}
\end{figure*}

\subsection{Overview}
\Cref{fig_methodology} illustrates the overall structure of DrUM for personalized generation in T2I diffusion models. DrUM customizes the T2I conditioning process using a transformer-based adapter to integrate historical prompts and the target prompt, enabling the synthesis of personalized outputs.

First, we identify the user profile from historical prompts and preference intensities (\eg, ratings) using a coreset sampling method (\Cref{sec_coreset}). This method effectively captures user preferences while reducing the number of references required for personalized conditioning. Next, to combine diverse user preferences into a personalized condition for T2I generation, we propose the \textbf{Pe}rsonalized \textbf{C}onditioning \textbf{A}dapter (PeCA), a transformer composed of only a few cross-attention layers for condition-level modeling (\Cref{sec_peca}). PeCA effectively integrates rich preference information with the target prompt in the latent space, overcoming constraints imposed by prompt-level limitations. Finally, we introduce a guidance mechanism (\Cref{sec_cond_guid}), which enables fine-grained control over preference intensity and personalization degree, ensuring highly refined personalized synthesis results.

These components are designed as adapter modules for open-source text encoders, making them applicable to popular foundation T2I models. Detailed explanations of these components are provided in the following sections.

\subsection{Coreset sampling}
\label{sec_coreset}

To effectively identify user profile, we sample key information from historical prompts. However, simple methods like random sampling may lead to information loss, and make it difficult to effectively utilize details such as preference intensity. Therefore, we employ the coreset sampling to exclude unnecessary references, reducing computational overhead while enhancing personalization.

Coreset sampling is a technique that selects a subset that effectively approximates the entire set \cite{agarwal2005geometric}. 
We perform sampling by measuring the CLIP similarity between historical prompts using the minimum enclosing ball technique. However, since precisely measuring distances between samples is NP-Hard \cite{korte2011combinatorial}, we adopt an iterative greedy approximation \cite{sener2018active} to reduce computational cost.

Our user profiling is summarized in \Cref{algo_coreset_sampling}. Initially, we randomly select $k$ from \(N\) samples to approximate the average score across entire embeddings \(E=\left(\textbf{e}_i\right)_{i=1}^N\) to reduce computational complexity. 
Subsequently, we sample core prompts incorporating the CLIP similarity \(\mathrm{Sim}_\textrm{CLIP}(*)\) (See \Cref{eq:clip_sim}) and preference intensities \(P=\left(p_i\right)_{i=1}^N\) to enable more refined user profiling.

\begin{algorithm}[b]
\caption{Coreset sampling}\label{algo_coreset_sampling}
\begin{algorithmic}[1]
\State \textbf{Input:} CLIP embeddings $E$, 
preference intensities $P$, sample size $n$, approximate size $k$
\State \textbf{Output:} Coreset indices $\mathcal{I}$
\State Initialize $\mathcal{I} \gets \varnothing$
\State Select subset \(E_k\subseteq E\) where \(\left|E_k\right|=k\)
\State Define operator \(\mathrm{Sim}(\mathbf{e}, E) = (\mathrm{Sim}_\textrm{CLIP}\left(\mathbf{e}, \mathbf{e}_i, p_i\right))_{i=1}^N \)
\State Distances \(D \gets \frac{1}{k} \sum_{\mathbf{e}\in E_k} \mathrm{Sim}(\mathbf{e}, E)\)

\Repeat 
    \State $s \gets \arg\max_i D$
    \State $\mathcal{I} \gets \mathcal{I} \cup \{s\}$
    \State $D \gets \min\left(D, \mathrm{Sim}(\mathbf{e}_s, E) \right)$
    \State $D_s \gets -\infty$
\Until{\(\left|\mathcal{I}\right|=n\)}
\end{algorithmic}
\end{algorithm}

\subsection{Condition-level modeling}
\label{sec_peca}

This section proposes a condition-level modeling approach that effectively integrates sufficient reference prompts in the latent space to overcome prompt-level limitations. However, it presents several challenges that must be addressed. First, there is no standard dataset or training method for constructing personalization based on condition embeddings. In T2I diffusion models, mapping users' historical data to the corresponding personalized data for training is a complex task. Moreover, personalized conditions must preserve the original embedding distribution to prevent performance degradation of foundation T2I models. Second, the positions and distributions of special or valid tokens can vary depending on the prompts. This difference affects class embeddings in text encoders like OpenCLIP, making token-level modeling more difficult. Finally, a generalizable solution is required, as different T2I models integrate text encoders into T2I conditioning in varying ways.

To achieve this, we propose PeCA, a transformer-based adapter attached with text encoders such as OpenCLIP and Google T5. PeCA generates personalized conditions by refining them using conditions tailored to each T2I model. PeCA is trained to reconstruct the target condition $y$, which helps the model learn reproducibility from the given conditions. This enables PeCA to effectively capture meaningful information at the token-level and reproduce it using cross-attention layers, where $y$ serves as both the key and value. We use cosine similarity as a training objective to keep the distribution of the reconstructed condition similar to that of the original text encoder while maintaining generative capability. From a technical perspective, $y$ is obtained from the embeddings before applying normalization, and PeCA on Google T5 add a single linear layer both before and after the adapter to handle the large embedding dimension.

\subsection{Conditioning guidance}
\label{sec_cond_guid}
PeCA effectively combines input conditions by concatenating them at the condition-level.
However, since softmax is applied only once along the token axis, this limits its ability to capture individual preferences precisely. Therefore, DrUM employs a guidance mechanism that enables detailed control over both preference intensity and personalization. 
This ensures that the generated outputs reflect user profiles accurately. To achieve this, the computation of attention scores for multiple reference conditions \(\left\{y_i\right\}_{i=1}^n\) and the target condition \(y_{n+1}\) is {decoupled} within the multi-head cross-attention module. Next, softmax is applied independently to each condition, and then overall personalization is modulated by a weighting parameter $\alpha$. 
Formally, the guidance formula is as follows:
\begin{equation}
w_i =
\begin{cases} 
(1 - \alpha) \cdot \text{Softmax}(s_i), & \text{if } i \in \text{target}, \\
\alpha \cdot \frac{\text{Softmax}(s_i) \cdot p_i}{\sum_j \text{Softmax}(s_j) \cdot p_j}, & \text{if } i \in \text{references}.
\end{cases}
\end{equation}
Here, \(w_i\) is the final attention weight for the index \(i\), and \(s_i\) is calculated from the scaled dot product of the query and the key embeddings.
Note that this mechanism also effectively accommodates cases that require separate class embedding (\eg, OpenCLIP).

Additionally, inspired by classifier-free guidance \cite{ho2022classifier}, our method uses unconditional text embeddings as a input query to establish the central distribution of text encoders. This approach ensures consistent and precise personalization by applying weight-based adjustments to each condition.
\section{Experiments}
This section describes the training settings for DrUM, outlines the evaluation setup and the baselines, and demonstrates effectiveness through various observations.

\begin{table}
\centering
\resizebox{0.8\columnwidth}{!}{
\begin{tabular}{ccccc}
\toprule
\multirow{4}{*}{T2I Models} & \multicolumn{4}{c}{Text Encoders} \\ \cmidrule(l){2-5} 
 & \multicolumn{3}{c}{OpenCLIP} & Google \\ \cmidrule(l){2-4}
 & ViT-L & ViT-H & ViT-bigG & T5 \\ \midrule
Stable Diffusion V1 & \checkmark &  &  & \\ 
Stable Diffusion V2 &  & \checkmark &  & \\ 
Stable Diffusion XL & \checkmark &  & \checkmark & \\ 
Stable Diffusion V3 & \checkmark &  & \checkmark & \checkmark \\ 
Flux & \checkmark &  &  & \checkmark \\ \bottomrule
\end{tabular}
}
\caption{Text encoders used in popular foundation T2I models.}
\label{tb_text_enc}
\end{table}

\subsection{Training}
We train DrUM using the Conceptual Captions 3M (CC3M) dataset \cite{sharma2018conceptual}. CC3M consists of 3.3 million images with captions covering diverse concepts (\eg, nature, animals, and objects). Since our personalized generation does not require images, we utilize only the set of prompts from CC3M.

We also focused on four encoders associated with popular foundation T2I models, as listed in \Cref{tb_text_enc}, despite DrUM's ability to integrate with most open-source text encoders. Using the same text encoders in the T2I models allows DrUM to scale seamlessly without training costs.

For training DrUM, we use 10 cross-attention layers on a single Nvidia RTX 6000 Ada GPU with a batch size of 256, initialized the learning rate at $5 \times 10^{-4}$, and employed the AdamW optimizer with a cosine annealing schedule over approximately 23K steps. For Google T5, we applied an encoding ratio of 4 to a linear layer both before and after PeCA. Note that we deliberately excluded the preference intensities $P$ and personalization degree $\alpha$, encouraging DrUM to precisely reconstruct the target condition while aligning with the original text encoder distribution.

\subsection{Evaluation setup}
We conduct both quantitative and qualitative evaluations on two datasets introduced in previous studies \cite{chen2024tailored, shen2024pmg}: (i) Personalized Image Prompt\footnote{https://github.com/zzjchen/Tailored-Visions/} (PIP): A large-scale dataset containing 3,115 users, each having 18 to $\sim$4,700 historical prompts, along with images generated by Stable Diffusion V1. (ii) MovieLens Latest Small\footnote{https://grouplens.org/datasets/movielens/} (ML): A dataset containing movie-watching records from 610 users, each having 20 to $\sim$2,600 interactions/ratings. We convert each movie information (\eg, title, genre, keyword, and description) into prompts using TMDB\footnote{https://www.themoviedb.org/} API. Notably, we use only the prompts for our evaluation, as the dataset images are not directly associated with the results generated by various T2I models from the given prompts, Additionally, we select the two most recent historical entries for each user, following the process in \cite{chen2024tailored}.

For quantitative evaluation, we measure the CLIP score and text alignment (Text align) between the personalized results and prompts by using the CLIP similarity \(\mathrm{Sim}_\textrm{CLIP}(*)\):
\begin{equation}
\mathrm{Sim}_\textrm{CLIP}\left(\mathbf{e}, \mathbf{r}_i, p_i\right) = \frac{\mathbf{e} \cdot \mathbf{r}_i}{\|\mathbf{e}\| \|\mathbf{r}_i\|} \cdot p_i 
\label{eq:clip_sim}
\end{equation}
where $\mathbf{e}$ is a CLIP embedding, $\mathbf{r}_i$ is the CLIP embedding for index $i$, and $p_i$ is the corresponding preference intensity.
The CLIP score and Text align are set by defining \(\mathbf{e}\) as a personalized generated image and a personalized T2I condition, respectively. Both metrics are averaged over target and historical text prompts \(\left\{\textbf{r}_i\right\}\).

For foundation T2I models, we use Stable Diffusion V1/V2/XL/V3 for quantitative evaluation, while qualitative case studies are conducted with Stable Diffusion V3 and FLUX. In terms of Text align, we employ only OpenCLIP text encoders ViT-L/H/bigG as they contain class tokens. Notably, the sampling ratio for user profiles was set to 10\%. The personalization degree $\alpha$ was set to 0.3 in PIP and 0.1 in ML for ensuring the target's originality, and the preference intensities (ratings) in ML were only applied as weights in the guidance mechanism and metrics.

\begin{table}[t]
\centering
\setlength\tabcolsep{4.5pt}
\begin{subfigure}{1\columnwidth}{
\resizebox{1\textwidth}{!}{
\begin{tabular}{clcccccc}
\toprule
\multicolumn{2}{c}{\multirow{2}{*}{CLIP score $\uparrow$}} & \multicolumn{3}{c}{PIP} & \multicolumn{3}{c}{ML} \\ \cmidrule(l){3-5} \cmidrule(l){6-8} 
\multicolumn{2}{l}{} & Target & History & Imp & Target & History & Imp \\ \midrule
\multirow{6}{*}{\shortstack{Stable \\ Diffusion \\ V1}} & - & 20.52 & 10.89 & - & 31.27 & 12.27 & - \\ \cmidrule(l){2-2} \cmidrule(l){3-5} \cmidrule(l){6-8} 
 & FABRIC & 24.33 & 15.85 & \tcbhighmath[blue5 eqbox]{\textcolor{BlueViolet}{\small{\textbf{+32.04\%}}}} & \textbf{30.80} & 12.13 & \tcbhighmath[red1 eqbox]{\textcolor{RedViolet}{\small{\textbf{-1.35\%}}}} \\ 
 & TV & 22.62 & 14.65 & \tcbhighmath[blue4 eqbox]{\textcolor{BlueViolet}{\small{\textbf{+22.40\%}}}} & 30.02 & 11.95 & \tcbhighmath[red1 eqbox]{\textcolor{RedViolet}{\small{\textbf{-3.31\%}}}} \\ 
 & PMG & - & - & - & 27.03 & 12.21 & \tcbhighmath[red2 eqbox]{\textcolor{RedViolet}{\small{\textbf{-7.03\%}}}} \\ 
 & DrUM & \textbf{25.13} & \textbf{16.03} & \tcbhighmath[blue5 eqbox]{\textcolor{BlueViolet}{\small{\textbf{+34.83\%}}}} & 29.49 & \textbf{13.07} & \tcbhighmath[blue1 eqbox]{\textcolor{BlueViolet}{\small{\textbf{+0.38\%}}}} \\ \cmidrule(l){1-2} \cmidrule(l){3-5} \cmidrule(l){6-8}
\multirow{6}{*}{\shortstack{Stable \\ Diffusion \\ V2}} & - & 21.20 & 11.71 & - & 33.32 & 11.30 & - \\ \cmidrule(l){2-2} \cmidrule(l){3-5} \cmidrule(l){6-8} 
 & FABRIC & 23.67 & 16.33 & \tcbhighmath[blue4 eqbox]{\textcolor{BlueViolet}{\small{\textbf{+25.61\%}}}} & 25.64 & 10.27 & \tcbhighmath[red3 eqbox]{\textcolor{RedViolet}{\small{\textbf{-16.05\%}}}} \\ 
 & TV & 23.38 & 15.50 & \tcbhighmath[blue4 eqbox]{\textcolor{BlueViolet}{\small{\textbf{+21.34\%}}}} & 31.58 & 11.28 & \tcbhighmath[red1 eqbox]{\textcolor{RedViolet}{\small{\textbf{-2.68\%}}}} \\ 
 & PMG & - & - & - & 27.81 & 11.15 & \tcbhighmath[red2 eqbox]{\textcolor{RedViolet}{\small{\textbf{-8.93\%}}}} \\ 
 & DrUM & \textbf{25.82} & \textbf{17.27} & \tcbhighmath[blue5 eqbox]{\textcolor{BlueViolet}{\small{\textbf{+34.68\%}}}} & \textbf{32.25} & \textbf{11.96} & \tcbhighmath[blue2 eqbox]{\textcolor{BlueViolet}{\small{\textbf{+1.34\%}}}} \\ \cmidrule(l){1-2} \cmidrule(l){3-5} \cmidrule(l){6-8} 
\multirow{6}{*}{\shortstack{Stable \\ Diffusion \\ XL}} & - & 23.26 & 13.64 & - & 33.89 & 8.92 & - \\ \cmidrule(l){2-2} \cmidrule(l){3-5} \cmidrule(l){6-8} 
 & FABRIC & - & - & - & - & - & - \\ 
 & TV & 25.44 & 17.19 & \tcbhighmath[blue3 eqbox]{\textcolor{BlueViolet}{\small{\textbf{+17.71\%}}}} & \textbf{32.04} & 8.69 & \tcbhighmath[red1 eqbox]{\textcolor{RedViolet}{\small{\textbf{-3.97\%}}}} \\ 
 & PMG & - & - & - & 28.63 & 9.24 & \tcbhighmath[red2 eqbox]{\textcolor{RedViolet}{\small{\textbf{-5.95\%}}}} \\ 
 & DrUM & \textbf{26.77} & \textbf{18.16} & \tcbhighmath[blue4 eqbox]{\textcolor{BlueViolet}{\small{\textbf{+24.13\%}}}} & 29.12 & \textbf{10.93} & \tcbhighmath[blue1 eqbox]{\textcolor{BlueViolet}{\small{\textbf{+4.26\%}}}} \\ \cmidrule(l){1-2} \cmidrule(l){3-5} \cmidrule(l){6-8} 
\multirow{6}{*}{\shortstack{Stable \\ Diffusion \\ V3}} & - & 22.60 & 12.75 & - & 37.69 & 9.76 & - \\ \cmidrule(l){2-2} \cmidrule(l){3-5} \cmidrule(l){6-8} 
 & FABRIC & - & - & - & - & - & - \\ 
 & TV & 23.85 & 15.42 & \tcbhighmath[blue3 eqbox]{\textcolor{BlueViolet}{\small{\textbf{+13.24\%}}}} & \textbf{35.83} & 9.37 & \tcbhighmath[red1 eqbox]{\textcolor{RedViolet}{\small{\textbf{-4.47\%}}}} \\ 
 & PMG & - & - & - & 32.48 & 9.67 & \tcbhighmath[red2 eqbox]{\textcolor{RedViolet}{\small{\textbf{-7.39\%}}}} \\ 
 & DrUM & \textbf{25.14} & \textbf{15.87} & \tcbhighmath[blue3 eqbox]{\textcolor{BlueViolet}{\small{\textbf{+17.86\%}}}} & 30.17 & \textbf{10.61} & \tcbhighmath[red2 eqbox]{\textcolor{RedViolet}{\small{\textbf{-5.60\%}}}} \\ \bottomrule
\end{tabular}
}}
\caption{CLIP score}
\end{subfigure}

\vspace{0.3em}
\setlength\tabcolsep{4.5pt}
\begin{subfigure}{1\columnwidth}{
\resizebox{1\textwidth}{!}{
\begin{tabular}{clcccccc}
\toprule
\multicolumn{2}{c}{\multirow{2}{*}{Text align $\uparrow$}} & \multicolumn{3}{c}{PIP} & \multicolumn{3}{c}{ML} \\ \cmidrule(l){3-5} \cmidrule(l){6-8} 
\multicolumn{2}{l}{} & Target & History & Imp & Target & History & Imp \\ \midrule
\multirow{5}{*}{\shortstack{OpenCLIP \\ ViT-L}} & - & 100.00 & 52.60 & - & 100.00 & 34.16 & - \\ \cmidrule(l){2-2} \cmidrule(l){3-5} \cmidrule(l){6-8} 
& TV & 68.50 & 44.65 & \tcbhighmath[red4 eqbox]{\textcolor{RedViolet}{\small{\textbf{-23.32\%}}}}& 86.14 & 30.41 & \tcbhighmath[red3 eqbox]{\textcolor{RedViolet}{\small{\textbf{-12.42\%}}}} \\ 
& PMG & - & - & - & 67.75 & 29.00 & \tcbhighmath[red4 eqbox]{\textcolor{RedViolet}{\small{\textbf{-23.67\%}}}} \\ 
& DrUM & \textbf{96.42} & \textbf{61.24} & \tcbhighmath[blue2 eqbox]{\textcolor{BlueViolet}{\small{\textbf{+6.42\%}}}} & \textbf{99.48} & \textbf{36.48} & \tcbhighmath[blue1 eqbox]{\textcolor{BlueViolet}{\small{\textbf{+3.13\%}}}} \\ 
\cmidrule(l){1-2} \cmidrule(l){3-5} \cmidrule(l){6-8} 
\multirow{5}{*}{\shortstack{OpenCLIP \\ ViT-H}} & - & 100.00 & 42.10 & - & 100.00 & 36.05 & - \\ \cmidrule(l){2-2} \cmidrule(l){3-5} \cmidrule(l){6-8} 
& TV & 64.79 & 37.86 & \tcbhighmath[red4 eqbox]{\textcolor{RedViolet}{\small{\textbf{-22.63\%}}}} & 85.88 & 33.93 & \tcbhighmath[red3 eqbox]{\textcolor{RedViolet}{\small{\textbf{-10.01\%}}}} \\ 
& PMG & - & - & - & 64.72 & 29.94 & \tcbhighmath[red4 eqbox]{\textcolor{RedViolet}{\small{\textbf{-26.12\%}}}} \\ 
& DrUM & \textbf{95.91} & \textbf{50.81} & \tcbhighmath[blue2 eqbox]{\textcolor{BlueViolet}{\small{\textbf{+8.30\%}}}} & \textbf{99.56} & \textbf{38.21} & \tcbhighmath[blue1 eqbox]{\textcolor{BlueViolet}{\small{\textbf{+2.77\%}}}} \\ \cmidrule(l){1-2} \cmidrule(l){3-5} \cmidrule(l){6-8} 
\multirow{5}{*}{\shortstack{OpenCLIP \\ ViT-bigG}} & - & 100.00 & 47.56 & - & 100.00 & 32.90 & - \\ \cmidrule(l){2-2} \cmidrule(l){3-5} \cmidrule(l){6-8} 
& TV & 68.03 & 41.69 & \tcbhighmath[red4 eqbox]{\textcolor{RedViolet}{\small{\textbf{-22.15\%}}}} & 86.82 & 30.21 & \tcbhighmath[red3 eqbox]{\textcolor{RedViolet}{\small{\textbf{-10.68\%}}}} \\ 
& PMG & - & - & - & 69.66 & 29.07 & \tcbhighmath[red4 eqbox]{\textcolor{RedViolet}{\small{\textbf{-21.00\%}}}} \\ 
& DrUM & \textbf{96.41} & \textbf{55.85} & \tcbhighmath[blue2 eqbox]{\textcolor{BlueViolet}{\small{\textbf{+6.92\%}}}} & \textbf{99.58} & \textbf{34.91} & \tcbhighmath[blue1 eqbox]{\textcolor{BlueViolet}{\small{\textbf{+2.84\%}}}} \\
\bottomrule
\end{tabular}
}}
\caption{Text align}
\end{subfigure}
\caption{Overall performance of DrUM compared to baselines. Imp means the average performance improvement rate compared to original models, and \textbf{bold} indicates the highest performance.}
\label{tb_overall}
\end{table}

\subsection{Baselines}
As mentioned in \Cref{sec_rel_per} and \ref{sec_pre_per}, personalized generation should effectively consider individual preferences to provide both creativity and diversity with minimal user intervention. However, ViPer \cite{salehi2025viper} requires detailed user comments, while Pigeon \cite{xu2025personalized} relies on LaViT’s built-in LDM with reference images, complicating its general evaluation on PIP and ML. Therefore, we selected FABRIC \cite{von2023fabric}, TV \cite{chen2024tailored}, and PMG \cite{shen2024pmg} as baselines for evaluating various foundation T2I models. FABRIC's reference pairs were determined based on preference intensity, when unavailable, the latest image was used as a positive. TV implemented ChatGPT 4o instead of 3.5 due to version issue to rewrite prompts with more advanced insight.

\subsection{Quantitative analysis}

\customsubsubsection{Effectiveness comparison}
\Cref{tb_overall} presents the personalized generation performance across various T2I models compared to baselines. In particular, ML consists of complex and diverse product information, leading to lower performance compared to PIP. Overall, DrUM demonstrates superior performance. However, with Stable Diffusion V3 on ML, the CLIP score and Text align intersect. This can be semantically attributed to the accumulation of complex product information within the limited canvas of the image, as text alignment increases with enhanced generative capability. Moreover, since the ML dataset contains diverse movie information, even small adjustments can significantly impact the target performance across all baselines. Notably, the enhanced Text align indicates that DrUM effectively generates content that aligns with both the target and historical preferences, even when the CLIP score is similar.

\begin{figure}[t]
    \centering

    \includegraphics[width=0.98\columnwidth]{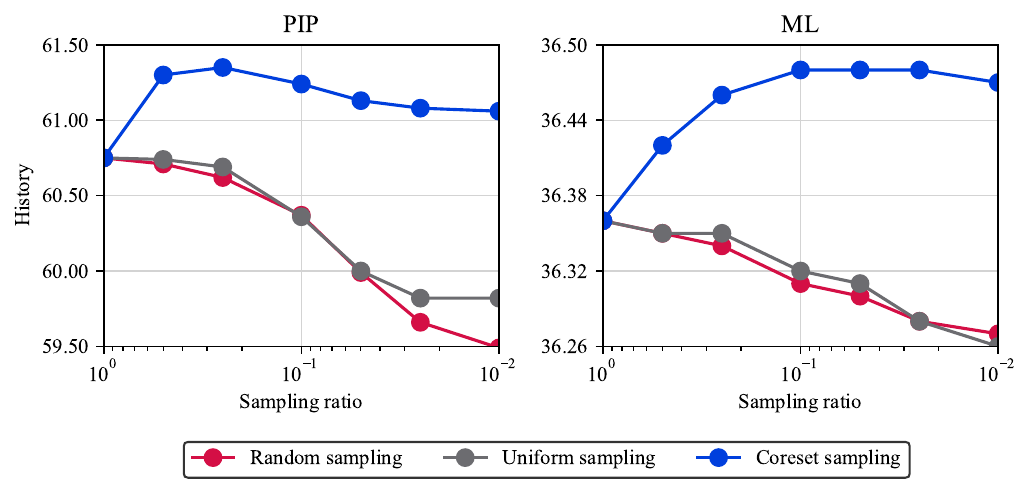}

    \vspace{-0.3em}

    \caption{Sampling performance of personalization on Text align for different methods using OpenCLIP ViT-L. Random selects randomly and uniform choices evenly.}
    \label{fig_sampling}
\end{figure}

\begin{table}[]
\centering
\resizebox{1\columnwidth}{!}{
\begin{tabular}{lcccccc}
\toprule
\multirow{2}{*}{CLIP score $\uparrow$} & \multicolumn{3}{c}{PIP} & \multicolumn{3}{c}{ML} \\ \cmidrule(l){2-4} \cmidrule(l){5-7} 
 & Target & History & Imp & Target & History & Imp \\ \midrule
- & 20.52 & 10.89 & - & 31.27 & 12.27 & - \\  \cmidrule(l){1-1} \cmidrule(l){2-4} \cmidrule(l){5-7} 
DrUM & 25.13 & 16.03 & \tcbhighmath[blue5 eqbox]{\textcolor{BlueViolet}{\small{\textbf{+34.83\%}}}} & 29.49 & 13.07 & \tcbhighmath[blue1 eqbox]{\textcolor{BlueViolet}{\small{\textbf{+0.38\%}}}} \\ 
w/o $S$ & \textbf{25.49} & 15.85 & \tcbhighmath[blue5 eqbox]{\textcolor{BlueViolet}{\small{\textbf{+34.88\%}}}} & \textbf{29.56} & 13.06 & \tcbhighmath[blue1 eqbox]{\textcolor{BlueViolet}{\small{\textbf{+0.46\%}}}} \\
w/o $G$ & 19.04 & 17.46 & \tcbhighmath[blue4 eqbox]{\textcolor{BlueViolet}{\small{\textbf{+26.55\%}}}} & 22.74 & 14.49 & \tcbhighmath[red1 eqbox]{\textcolor{RedViolet}{\small{\textbf{-4.64\%}}}}\\
w/o $S$ \& $G$ & 18.13 & \textbf{18.85} & \tcbhighmath[blue4 eqbox]{\textcolor{BlueViolet}{\small{\textbf{+30.71\%}}}} & 21.23 & \textbf{15.44} & \tcbhighmath[red1 eqbox]{\textcolor{RedViolet}{\small{\textbf{-3.15\%}}}} \\ \bottomrule
\end{tabular}
}

\caption{Ablation study on the effect of removing sampling $S$ and guidance mechanism $G$ on CLIP score using Stable Diffusion V1 and OpenCLIP ViT-L.}
\label{tb_abl_study}
\end{table}

\begin{figure*}[t]
    \centering
    \includegraphics[width=0.83\textwidth]{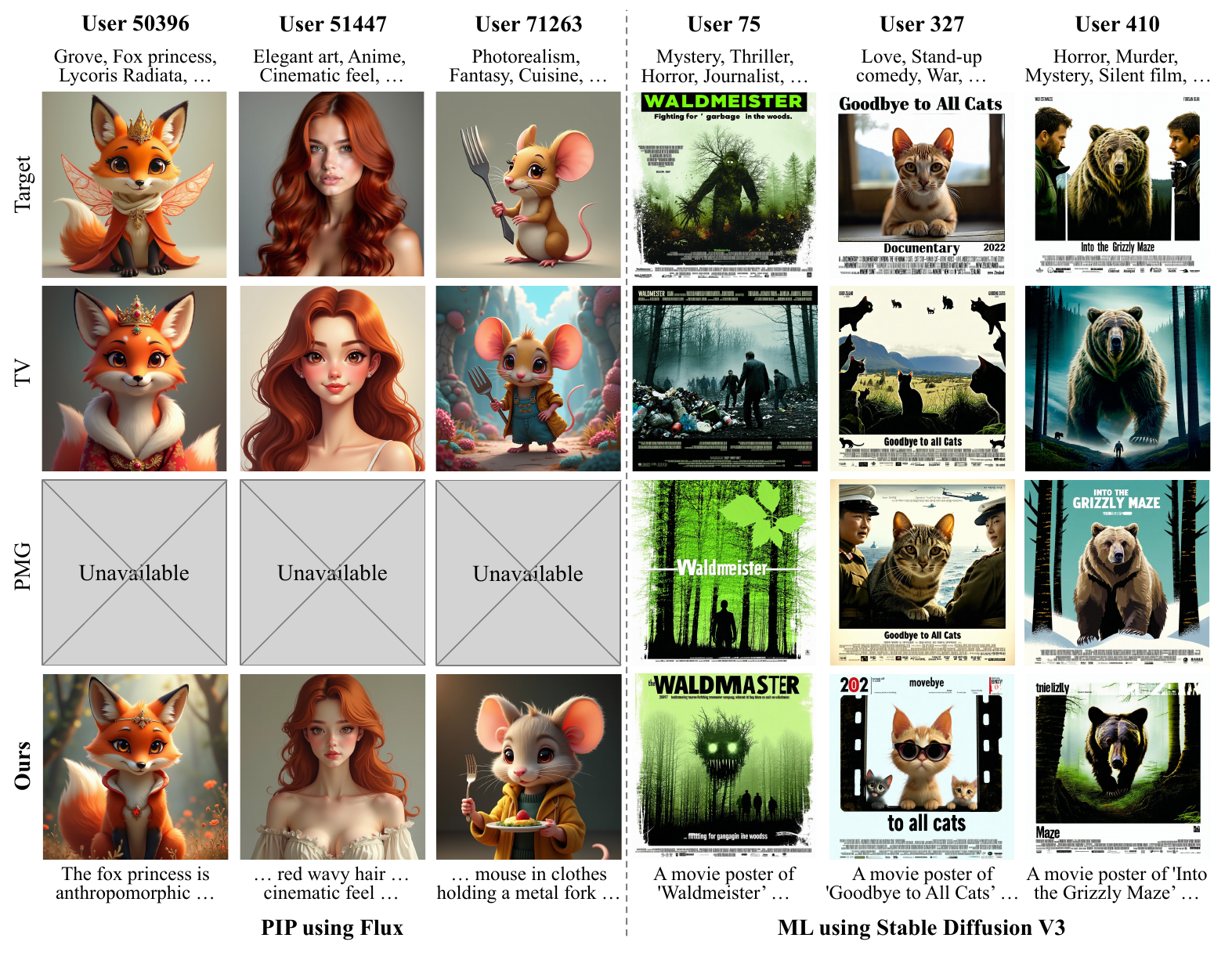}
    \vspace{-1.1em}
    \caption{Qualitative comparison using \textit{multiple prompts} with baselines. Key keywords extracted from historical user prompts.}
    \label{fig_comparison}
\end{figure*}

\begin{figure*}[t]
    \centering
    \vspace{-0.5em}
    \includegraphics[width=0.93\textwidth]{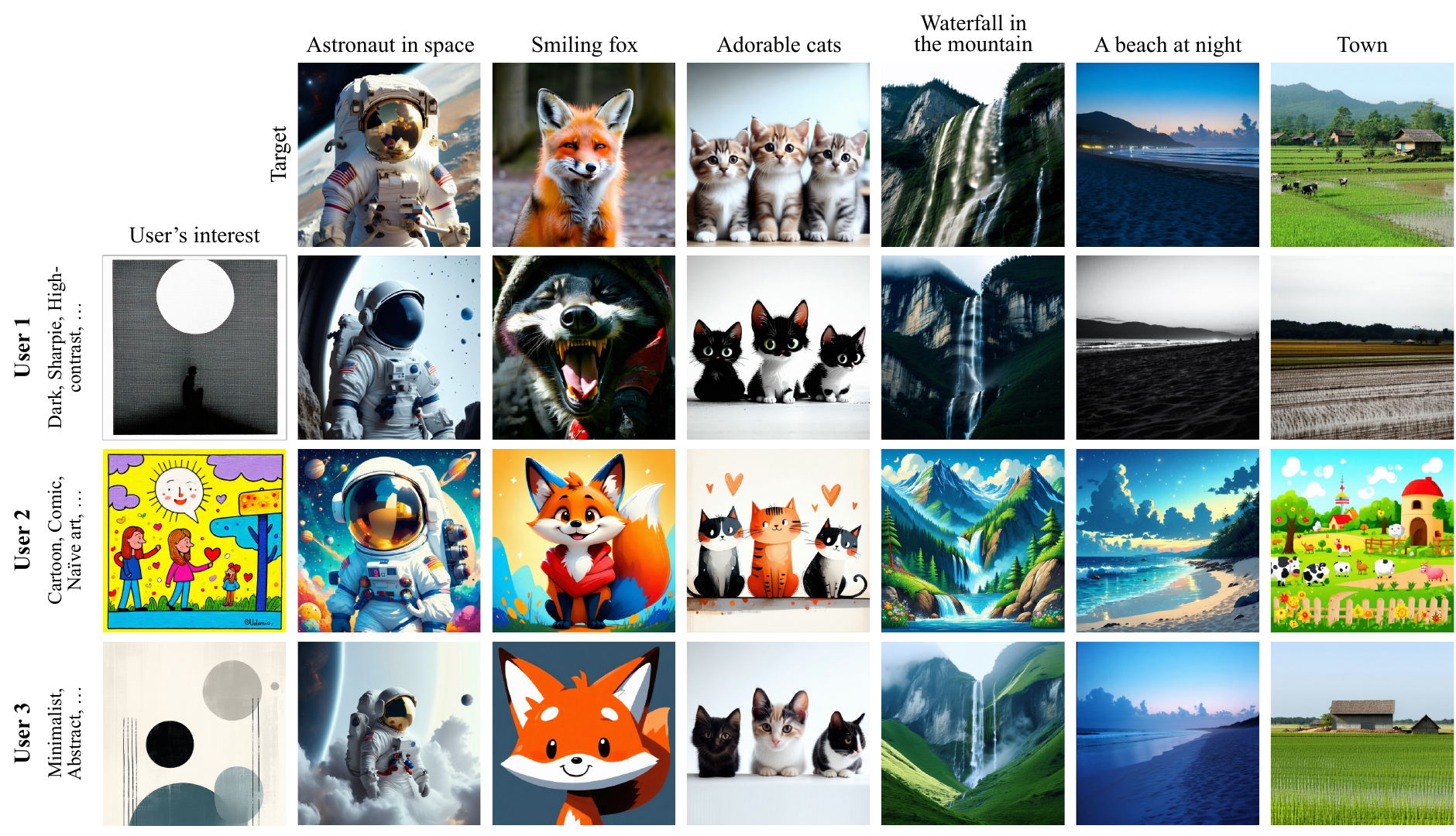}
    \vspace{-0.8em}
    \caption{Prompt-based style transfer using a \textit{single style prompt} on Stable Diffusion V3. User's interest illustrates the results from \textit{single style prompts}.}
    \label{fig_style}
\end{figure*}

\customsubsubsection{Impact of sampling method}
\Cref{fig_sampling} represents how different sampling methods affect personalization performance as the sampling ratio changes. Coreset sampling consistently achieves excellent performance across all ratios. At a 10\% sampling ratio, it shows superior overall performance, so we select it for user profiling. As a result, our coreset sampling method effectively and efficiently integrates more of each user's history into their profile.

\customsubsubsection{Ablation study}
\Cref{tb_abl_study} presents an ablation study on the effect of removing modules from DrUM. When only $G$ is removed, a significant drop in the target score is observed, highlighting the essential role of guidance mechanism in balancing between target and history. When both $S$ and $G$ are removed, DrUM shows a convergence of performance between target and history. This is likely due to an overabundance of historical information, which causing the results to lose a clear separation. Although removing only $S$ leads to slight performance changes, sampling is crucial in reducing computational costs and prioritizing the most relevant user information. Thus, DrUM not only enhances efficiency but also better preserves the balance between personalization and target fidelity.

\subsection{Case study}

\customsubsubsection{Qualitative comparison}
\Cref{fig_comparison} compares personalized content on Stable Diffusion V3 and Flux with the baselines. TV is a general approach but struggles to incorporate user preferences, and often distorts original content in ML. PMG captures key user preferences better than TV, but relies on keyword-centered modeling in ML, which lacks detail. In contrast, DrUM effectively preserves the originality while precisely reflecting user preferences. As shown in the quantitative results, it is visually demonstrated that higher Text align provides more effective personalization, even when the CLIP score is similar. Furthermore, for user 71263, Flux fails to generate 'clothes', whereas DrUM successfully satisfies this requirement while aligning with user preferences. This indicates that personalized generation not only reflects individual preferences but also enhances originality.

\begin{figure}[t]
    \centering
    \includegraphics[width=0.98\columnwidth]{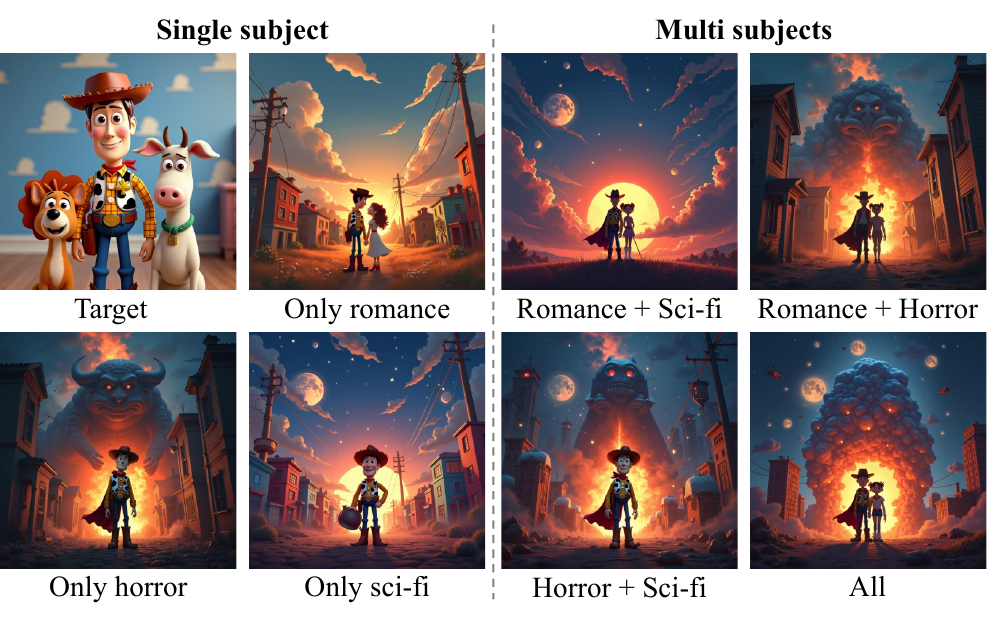}
    \vspace{-0.9em}
    \caption{Prompt-based subject transfer using \textit{multiple prompts} on Flux.}
    \label{fig_movie_flux}
    \vspace{-0.61em}
\end{figure}

\customsubsubsection{Real-world scenarios}
We introduce two additional scenarios for real-world users. \Cref{fig_style} shows prompt-based style transfer. Rather than constraining users to write detailed prompts within a limited token length, we recommend separating prompts into target and visual attributes. This approach enables a more comprehensive representation of target prompts without requiring detailed information such as style, texture. \Cref{fig_movie_flux} illustrates prompt-based multi-subject transfer. DrUM effectively captures both single and multiple subjects, while preserving their characteristics. Similar to the previous scenario, this is easily achieved by separating subject attributes in the prompt.

\begin{figure}[t]
    \centering
    \includegraphics[width=1\columnwidth]{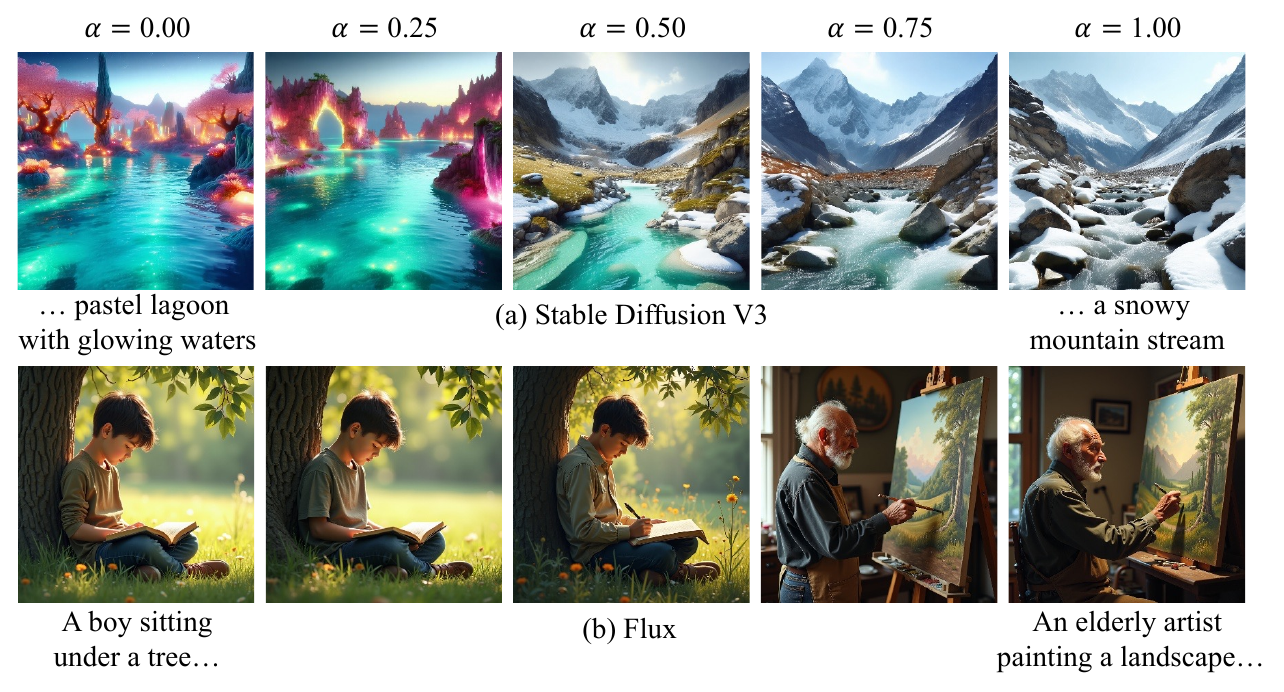}
    \vspace{-1.8em}
    \caption{Impact of personalization degree $\alpha$ on different prompts.}
    \label{fig_dgree}
    \vspace{-1.5em}
\end{figure}

\customsubsubsection{Personalization degree}
DrUM enables flexible adjustment of personalization degree at the latent-level. \Cref{fig_dgree} shows how personalized results change between two different prompts on the given degree. In \Cref{fig_dgree}(a), the background and season are semantically interpolated, while in \Cref{fig_dgree}(b), the occupation and age are blended. Even simple adjustments yield effective personalized results.

\subsection{Discussion}

In our experiments, DrUM showed a slight decrease in ML, likely due to the complexity of product information. Over-incorporating rich details can inadvertently compromise the output. In contrast, as shown in PIP, DrUM achieves very high performance with natural prompts commonly used by real-world users. This indicates that personalization should be applied cautiously to products or similarly complex descriptions. We also note that adjusting the number of references or refining token-level information may be beneficial, though this remains to be explored further.

On the other hand, our architecture employs only a small number of cross-attention layers, which may not fully capture all patterns. Our objective was to showcase the potential of direct modeling without relying on large-scale models and to assess the reproducibility of the embeddings derived through transformers. While our approach provides promising results, a more in-depth exploration with deeper architectures remains a subject for future research.
\section{Conclusion}
We propose DrUM, a novel method for personalized generation via condition-level modeling to precisely integrate individual user preferences into T2I diffusion models. Experiments demonstrate that DrUM's modules effectively generate high-quality personalized synthesis results, while ensuring creativity and diversity across various foundation T2I models, without additional fine-tuning. Future work will explore more optimized architectures and modeling approaches for complex inputs like product information.
\section*{Acknowledgements}
This work was supported in part by Institute of Information \& communications Technology Planning \& Evaluation (IITP) grants funded by the Korea government (MSIT) (No. 2022-0-00448/RS-2022-II220448, Deep Total Recall: Continual Learning for Human-Like Recall of Artificial Neural Networks, and No. RS-2022-00155915, Artificial Intelligence Convergence Innovation Human Resources Development (Inha University)), and in part by INHA UNIVERSITY Research Grant.
{
    \small
    \bibliographystyle{ieeenat_fullname}
    \bibliography{main}
}

\end{document}